\documentclass{article}

\usepackage{arxiv}

\usepackage[utf8]{inputenc} % allow utf-8 input
\usepackage[T1]{fontenc}    % use 8-bit T1 fonts
\usepackage{hyperref}       % hyperlinks
\usepackage{url}            % simple URL typesetting
\usepackage{booktabs}       % professional-quality tables
\usepackage{amsfonts}       % blackboard math symbols
\usepackage{nicefrac}       % compact symbols for 1/2, etc.
\usepackage{microtype}      % microtypography
\usepackage{lipsum}
\usepackage{graphicx}
\usepackage{amsmath}
\usepackage{amssymb}
\usepackage{mathtools}
\usepackage{amsthm}
\usepackage{multirow}
\title{Sparse Centroid-Encoder: A Nonlinear Model for Feature Selection}

\author{
 Tomojit Ghosh \\
  Department of Computer Science\\
  Colorado State University\\
  Fort Collins, CO 80523-1873 \\
  \texttt{tomojit.ghosh@colostate.edu} \\
   \And
 Michael Kirby \\
  Department of Mathematics\\
  Colorado State University\\
  Fort Collins, CO 80523-1873 \\
  \texttt{Kirby@math.colostate.edu} \\
}

\begin{document}
\maketitle
\begin{abstract}
%We develop a sparse optimization problem for the determination of the total set of features 
%that discriminate two or more classes. This is a sparse implementation of the centroid-encoder for nonlinear data reduction and visualization called Sparse Centroid-Encoder (SCE). We also provide a feature selection framework that first ranks each feature by its occurrence, and the optimal number of features is chosen using a validation set.
%The algorithm is applied to a wide variety of data sets including, single-cell biological data, high dimensional infectious disease data, hyperspectral data, image data, and speech data. We compared our method to various state-of-the-art feature selection techniques, including two neural network-based models (DFS, and LassoNet), Sparse SVM, and Random Forest. We empirically showed that SCE features produced better classification accuracy on the unseen test data, often with fewer features. 
Autoencoders have been widely used as a nonlinear tool for data dimensionality reduction. While autoencoders don't utilize the label information, Centroid-Encoders (CE)\cite{ghosh2022supervised} use the class label in their learning process. In this study, we propose a sparse optimization using the Centroid-Encoder architecture to determine a minimal set of features that discriminate between two or more classes. The resulting algorithm, Sparse Centroid-Encoder (SCE), extracts discriminatory features in groups using a sparsity inducing $\ell_1$-norm while mapping a point to its class centroid. One key attribute of SCE is that it can extract informative features from a multi-modal data set, i.e., data sets whose classes appear to have multiple clusters.
The algorithm is applied to a wide variety of real world data sets, including single-cell data, high dimensional biological data, image data, speech data, and accelerometer sensor data. We compared our method to various state-of-the-art feature selection techniques, including supervised Concrete Autoencoders (SCAE), Feature Selection Network (FsNet), deep feature selection (DFS), Stochastic Gate (STG), and LassoNet. We empirically showed that SCE features often produced better classification accuracy than other methods on sequester test set.
\end{abstract}

% keywords can be removed

\section{Introduction}
Technological advancement has made high-dimensional data readily available. For example, in bioinformatics, the researchers seek to understand the gene expression level with microarray or next-generation sequencing techniques where each point consists of over 50,000 measurements \cite{Pease5022,shalon1996dna,metzker2010sequencing,reuter2015high}. 
%The high-resolution imaging technology has produced digital images with many pixels in them. A 256 x 256 colored digital image amounts to 3x256x256=196608-dimensional vector. Often the features in a high-dimensional data set are noisy, redundant, and irrelevant (\cite{alelyani2018feature}), which degrades the performance of a Machine Learning task (\cite{raj2020efficient}). 
The abundance of features demands the development of feature selection algorithms to improve a Machine Learning task, e.g., classification. Another important aspect of feature selection is knowledge discovery from data. Which biomarkers are important to characterize a biological process, e.g., the immune response to infection by respiratory viruses such as influenza \cite{o2013iterative}? Additional benefits of feature selection include improved visualization and understanding of data, reducing storage requirements, and faster algorithm training times.

Feature selection can be accomplished in various ways that can be broadly categorized into the filter, wrapper, and embedded methods. In a filter method, each variable is ordered based on a score. After that, a threshold is used to select the relevant features \cite{lazar2012survey}. Variables are usually ranked using correlation \cite{guyon2003introduction,yu2003feature}, and mutual information \cite{vergara2014review,fleuret2004fast}. In contrast, a wrapper method uses a model and determines the importance of a feature or a group of features by the generalization performance of the predetermined model \cite{el2016review, hsu2002annigma}. Since evaluating every possible combination of features becomes an NP-hard problem, heuristics are used to find a subset of features. Wrapper methods are computationally intensive for larger data sets, in which case search techniques like Genetic Algorithm (GA) \cite{goldberg1988genetic} or Particle Swarm Optimization (PSO) \cite{kennedy1995particle} are used. In embedded methods, feature selection criteria are incorporated within the model, i.e., the variables are picked during the training process \cite{lal2006embedded}. Iterative Feature Removal (IFR) uses the absolute weight of a Sparse SVM model as a criterion to extract features from the high dimensional biological data set \cite{o2013iterative}.

Mathematically feature selection problem can be posed as an optimization problem on $\ell_0$-norm, i.e., how many predictors are required for a machine learning task. As the minimization of $\ell_0$ is intractable (non-convex and non-differentiable), $\ell_1$-norm is used instead, which is a convex proxy of $\ell_0$ \cite{tibshirani1996regression}. Note $\ell_1$ is not differentiable at 0, but the problem can be tackled by leveraging on sub-gradient methods \cite{boyd2003subgradient,duchi2011adaptive}. Since the introduction of these seminal papers \cite{candes2006stable,candes2005decoding}, the use of 1-norm has surged significantly in research work. Despite some difficulties \cite{zou2006adaptive,zou2005regularization}, the $\ell_1$ has been used for the feature selection task in linear \cite{fonti2017feature,muthukrishnan2016lasso,kim2004gradient,o2013iterative,chepushtanova2014band} as well as in nonlinear regime \cite{li2016deep,scardapane2017group,li2020efficacy}.

This paper proposes a new embedded variable selection approach called Sparse Centroid-Encoder (SCE) to extract features when class labels are available. Our method extends the Centroid-Encoder model \cite{GHOSH201826,GhKi2020}, where we applied a $l_1$ penalty to a sparsity promoting layer between the input and the first hidden layer. We evaluate this Sparse Centroid-Encoder on diverse data sets and show that the selected features produce better generalization than other state-of-the-art techniques. Our results showed that SCE picked fewer features to obtain high classification accuracy. As a feature selection tool, SCE uses a single model for the multi-class problem without the need to create multiple one-against-one binary models typical of linear methods, e.g., Lasso \cite{tibshirani1996regression}, or Sparse SVM \cite{chepushtanova2014band}. Although SCE can be used both in binary and multi-class problems, we focused on the multi-class feature selection problem in this paper. The work of \cite{li2016deep} also uses a similar sparse layer between the input and the first hidden with an Elastic net penalty while minimizing the classification error with a softmax layer.
The authors used Theano's symbolic differentiation \cite{bergstra2010theano} to impose sparsity.
%on the sparse layer. 
In contrast, our approach minimizes the centroid-encoder loss with an explicit differentiation of the $l_1$ function using the sub-gradient. Unlike DFS, our model can capture the intra-class variability by using multiple centroids per class. This property is beneficial for multi-modal data sets.

The article is organized as follows: 
In Section \ref{SCE} we present the Sparse Centroid-Encoder algorithm.
In Section \ref{experiments} we apply SCE to a range of bench-marking
data sets taken from the literature.
In Section \ref{lit}, we review related work, for both linear and
non-linear feature selection.
In Section \ref{dis_cons}, we present our discussion and conclusion.

\section{Sparse Centroid-Encoder }
\label{SCE}
Centroid-encoder (CE) neural networks are the starting point of our approach~\cite{GhKi2020,GHOSH201826,aminian2021early}. We present a brief overview of CEs and demonstrate 
%First, we briefly describe centroid-encoders and then show 
how they can be extended to perform non-linear feature selection.

\subsection{Centroid-encoder}
 The CE neural network is a variation of an autoencoder and can be used for both visualization and classification tasks. Consider a data set with $N$ samples and $M$ classes. The classes denoted $C_j, j = 1, \dots, M$ where the indices of the data associated with class $C_j$ are denoted $I_j$. We define centroid of each class as 
$c_j=\frac{1}{|C_j|}\sum_{i \in I_j} x^i$ where $|C_j|$ is the cardinality of  class $C_j$. Unlike autoencoder, which maps each point $x^i$ to itself, the CE maps each point $x^i$  to its class centroid $c_j$ 
by minimizing the following cost function over the parameter set $\theta$:
\begin{equation}
\begin{aligned}
 \mathcal{L}_{ce}(\theta)=\frac{1}{2N}\sum^M_{j=1} \sum_{i \in I_j}\|c_j-f(x^i; \theta))\|^2_2
 \label{equation:CECostFunction}
\end{aligned}
\end{equation}
	
The mapping $f$ is composed of a dimension reducing mapping
$g$ (encoder) followed by a dimension increasing reconstruction
mapping $h$ (decoder). The output of the encoder is used as a supervised visualization tool \cite{GhKi2020,GHOSH201826}, and attaching another layer to map to the one-hot encoded labels performs robust classification \cite{aminian2021early}.

\subsection{Sparse Centroid-encoder for feature selection}
\begin{figure}[t!]
    %\vspace{-0.5cm}
	\centering
	\includegraphics[width=10.0cm,height=16.cm]{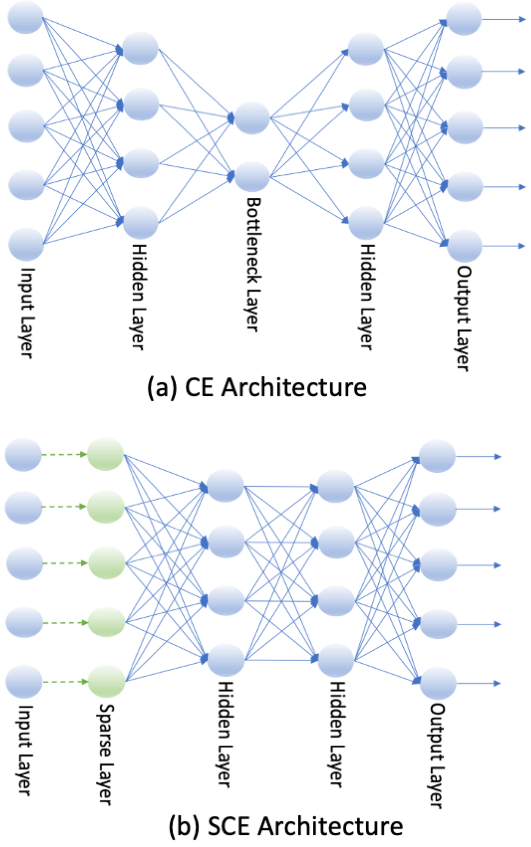}
	\caption{The architecture of Centroid-encoder and Sparse Centroid-encoder. Notice the Centroid-encoder uses a bottleneck architecture which is helpful for visualization. In contrast, the Sparse Centroid-encoder doesn't use any bottleneck architecture; instead, it employs a sparse layer between the input and the first hidden layer to promote feature sparsity.}
	\label{fig:SCE_arch}
\end{figure}

The Sparse Centroid-encoder (SCE) is a modification to the centroid-encoder architecture as shown in Figure \ref{fig:SCE_arch}. Unlike centroid-encoder, we haven't used a bottleneck architecture as visualization is not our aim here. The input layer is connected to the first hidden layer via the sparsity promoting layer (SPL). Each node of the input layer has a weighted one-to-one connection to each node of the SPL. 
The number of nodes in these two layer are the same. The nodes in SPL don't have any bias or non-linearity. The SPL is fully connected to the first hidden layer, therefore the weighted input from the SPL will be passed to the hidden layer in the same way that of a standard feed forward network. During training, a $l_{1}$ penalty will be applied to the weights connecting the input layer and SPL layer. The sparsity promoting $l_{1}$ penalty will drive most of the weights to near zero and the corresponding input nodes/features can be discarded. Therefore, the purpose of the SPL is to select important features from the original input. Note we only apply the $l_{1}$ penalty to the parameters of the SPL.

Denote $\theta_{spl}$ to be the parameters (weights) of the SPL and $\theta$ to be the parameters of the rest of the network. The cost function of sparse centroid-encoder is given by
%in Equation \ref{equation:SCECostFunction} 
\begin{equation}
\begin{aligned}
 \mathcal{L}_{sce}(\theta)=\frac{1}{2N}\sum^M_{j=1} \sum_{i \in I_j}\|c_j-f(x^i; \theta))\|^2_2 + \lambda \|\theta_{spl}\|_{1}
 \label{equation:SCECostFunction}
\end{aligned}
\end{equation}
where $\lambda$ is the hyper-parameter which controls the sparsity. A larger value of $\lambda$ will promote higher sparsity resulting more near-zero weights in SPL. In other words, $\lambda$  is a knob that controls the number of features selected from 
the input data.

Like centroid-encoder, we trained sparse centroid-encoder using error backpropagation, which requires the gradient of the cost function of Equation \ref{equation:SCECostFunction}. As $\ell_1$ function is not differentiable at $0$, we implement this term using the sub-gradient\cite{boyd2003subgradient}. We trained SCE using Scaled Conjugate Gradient Descent \cite{moller1993scaled} on the full training set. Like any neural network-based model, the hyperparameters of SCE need to be tuned for optimum performance. Table \ref{table:SCEHyperparameters} contains the list with the range of values we used in this paper. We used validation set to choose the optimal value. For a small sample size data set ( high-dimensional biological data), we ran a five-fold cross validation on the training set to pick the optimum value. The Python implementation of Sparse Centroid-Encoder will be provided in Supplementary Material.

\begin{table}[!ht]	
	\vspace{1mm}
	\centering
	\begin{tabular} {|c|c|}
		% start header
		\hline 	% Makes 2 fancy lines
		Hyper parameter & Range of Values \\
		\hline
		\#SCG Iteration & \{25,50,75,100\} \\
		\hline
		\# Hidden Layers (L) & \{1, 2\} \\
		\hline
        \# Hidden Nodes (H) & \{50,100,200,250,500\} \\
        \hline
        Activation Function & Hyperbolic tangent (tanh) \\
        \hline
        $\lambda$ & \{0.01, 0.001, 0.0001, 0.0002, 0.0004, 0.0006, 0.0008\} \\
        \hline
        \# Center/Class & \{1, 2, 3, 4, 5\} \\
        \hline
	\end{tabular}	
	\caption{Hyperparameters for Sparse Centroid-Encoder.}
	\label{table:SCEHyperparameters}
\end{table}

\subsubsection{Feature Cut-off}
\label{feature_cut-off}
The 1-norm of the sparse layer (SPL) drives a lot of weight to near zero. Often hard thresholding or a ratio of two consecutive weights is used to pick the nonzero weight \cite{o2013iterative}. We take a different approach. After training SCE, we arrange the absolute value of the weights of the SPL in descending order. And then find the elbow of the curve. We measure the distance of each point of the curve to the straight line formed by joining the first and last points of the curve. The point with the largest distance is the position (P) of the elbow. We pick all the features whose absolute weight is greater than that of P; see panel (c), (d) of Figure\ref{fig:SCE_analysis} and panel (b), (c) of Figure \ref{fig:Feature_Stability}.

\subsection{Empirical Analysis of SCE}
\label{sparsity_analysis}
\begin{figure}[b!]
    \centering
    \includegraphics[width=12.5cm, height=10.5cm]{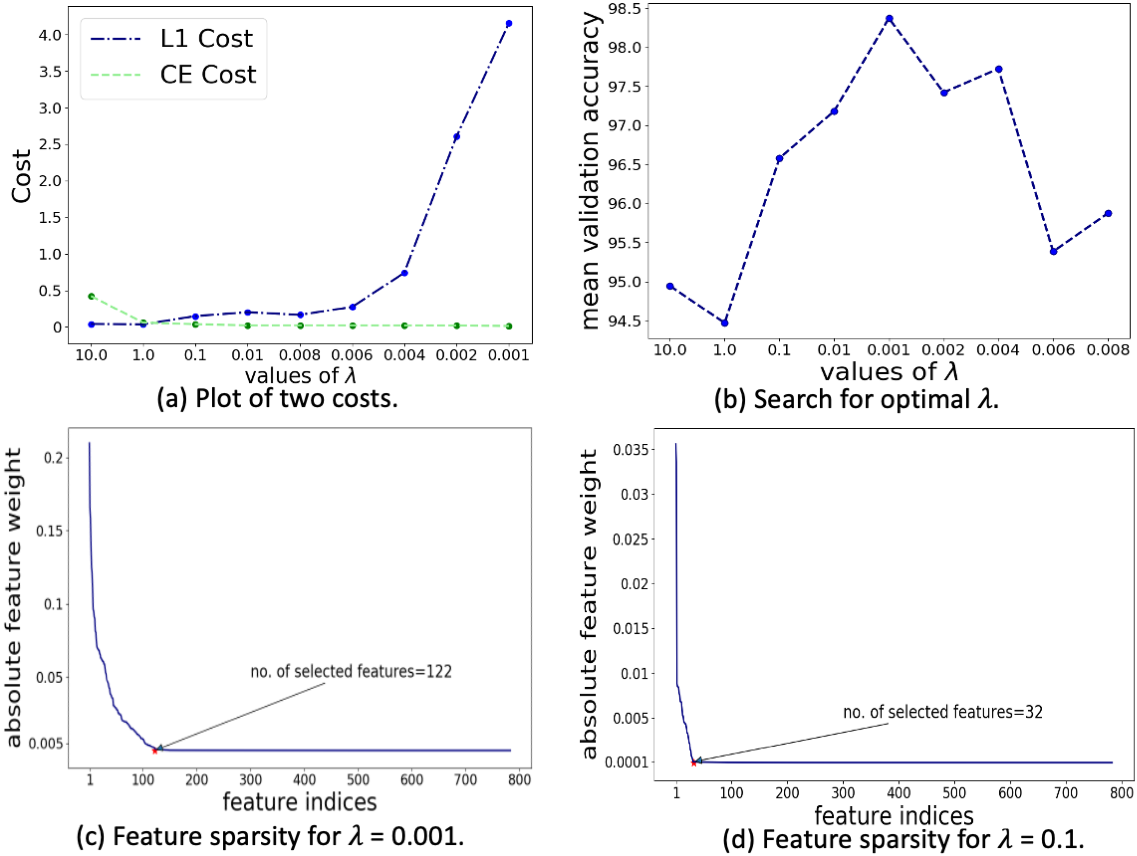}
    %\caption{Analysis of Sparse Centroid-encoder}
    \label{fig:GSE73072_Sparsity}
    \caption{Analysis of Sparse Centroid-encoder. (a) Change of the two costs over $\lambda$. (b) Change of validation accuracy over $\lambda$. (c) Sparsity plot of the weight of $W_{SPL}$ for $\lambda = 0.001$. (d) Same as (c) but $\lambda = 0.1$.}
    \label{fig:SCE_analysis}
\end{figure}

In this section we present an empirical analysis of our model. The results of feature selection for the digits 5 and 6 from the MNIST set are displayed in Figure \ref{fig:SCE_analysis}. In panel (a), we compare the two terms
that contribute to Equation \ref{equation:SCECostFunction}, i.e., the 
centroid-encoder and $l_{1}$ costs,
%, i.e., the sum of the absolute values of weights of the sparse layer 
 weighted with different values of $\lambda$.  As expected, we observe that the CE cost monotonically decreases with 
 %the decrease of
 $\lambda$, while the  $l_{1}$ cost increases as 
 $\lambda$ decreases. %The change of these costs is expected.
 For larger values of $\lambda$, the model focuses more on minimizing the $l_{1}$-norm of the sparse layer, which results in smaller values. In contrast, the model pays more attention to minimizing the CE cost for small $\lambda$s; hence we notice smaller CE cost and higher $l_{1}$ cost.

\begin{figure}[h!]
    %\vspace{-0.5cm}
	\centering
	\includegraphics[width=13.25cm,height=12.5cm]{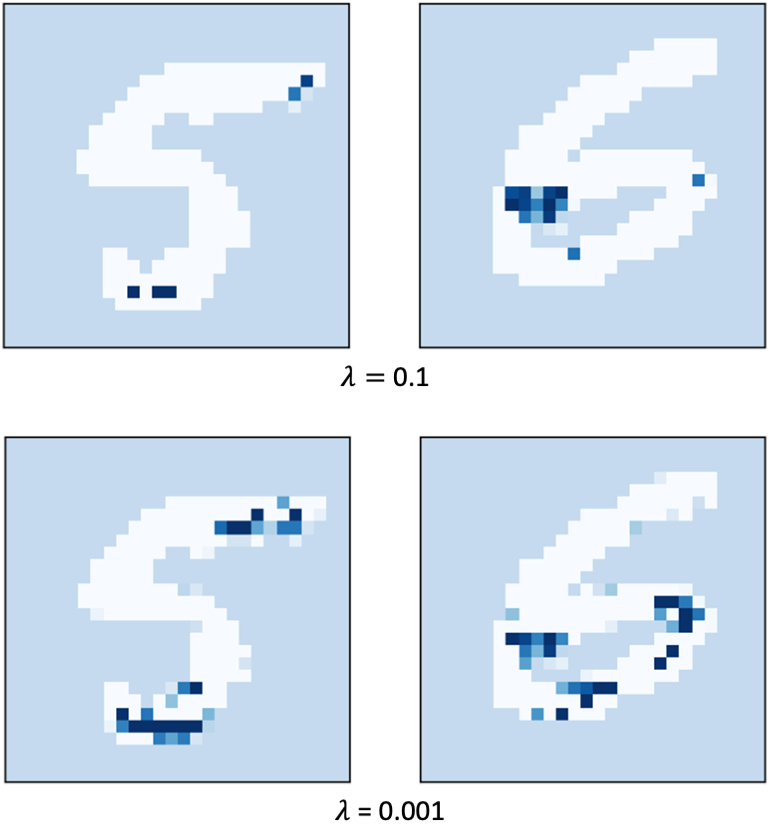}
	\caption{Demonstration of the sparsity of Sparse Centroid-encoder on MNIST digits 5 and 6. The digits are shown in white, and the selected pixels are marked using blue—the darkness of blue indicates the relative importance of the pixel to distinguish the two digits. We showed the selected pixels for two choices of $\lambda$. Notice that for $\lambda = 0.1$, the model chose the lesser number of features, whereas it picked more pixels for $\lambda = 0.001$. $\lambda$ is the nob which controls the sparsity of the model.}
	\label{fig:SCE_Sparsity}
\end{figure}

Panel (b) of Figure \ref{fig:SCE_analysis} shows the accuracy on a validation set as a function nine different values of $\lambda$; the validation accuracy reached its peak for $\lambda = 0.001$. 
%We chose $\lambda$ using a validation set for all the data sets. 
In panels (c) and (d), we plotted the magnitude of the feature weights of the sparse layer in descending order.
The sharp decrease in the magnitude of the weights
demonstrates the promotion of sparsity by SCE. The model
effectively ignores features by setting their weight to approximately zero. Notice the model produced a sparser solution for $\lambda = 0.1$, selecting only 32 features compared to 122 chosen variables for $\lambda = 0.001$. Figure \ref {fig:SCE_Sparsity} shows the position of the selected features, i.e.,  pixels, on the digits 5 and 6. The intensity of the color represents the feature's importance. Dark blue signifies a higher absolute value of the weight, whereas light blue means a smaller absolute weight.

\begin{figure}[h!]
	\centering
	\includegraphics[width=14.5cm,height=13.5cm]{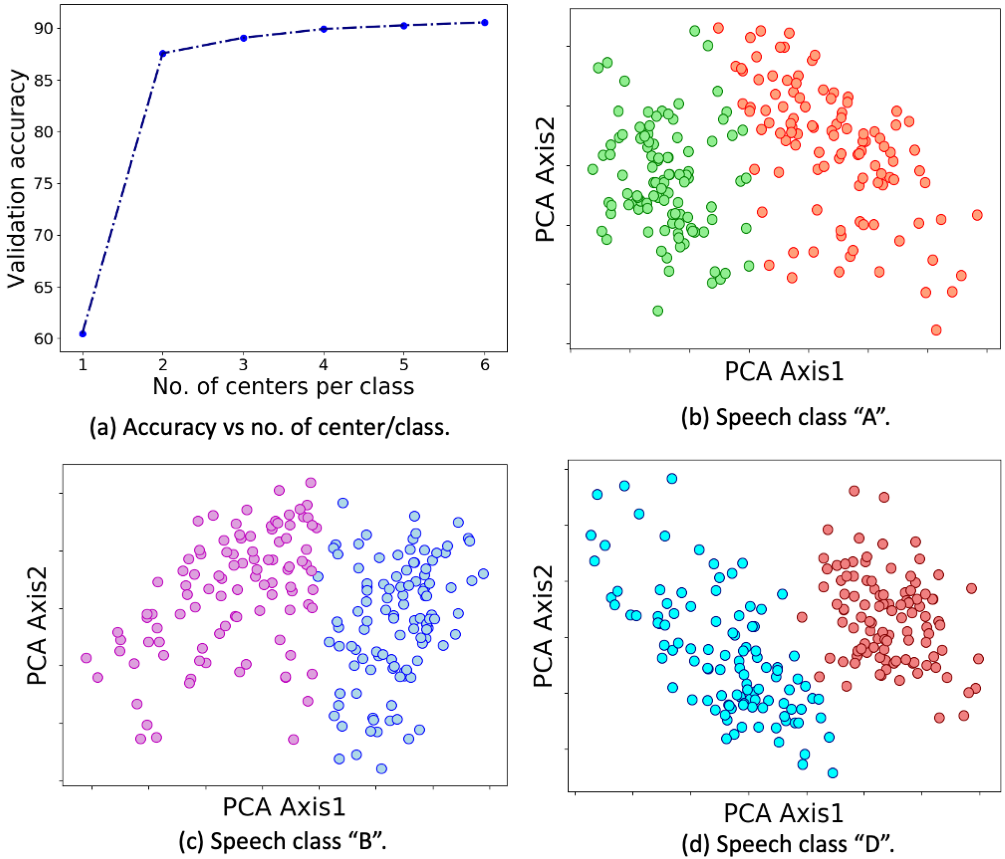}
	\caption{Sparse Centroid-encoder for multi-modal data set. Panel (a) shows the increase in validation accuracy over the number of centroids per class. Panel (b)-(d) shows the two-dimensional PCA plot of the three speech classes.}
	\label{fig:Isolet_Analysis}
\end{figure}

Our next analysis shows how SCE  extracts informative features from a multi-modal dataset, i.e., data sets whose classes appear to have multiple clusters.  In this case, one center per class may not be optimal, e.g., ISOLET data. To this end, we trained SCE using a different number of centers per class where the centers were determined using standard $k$-Means algorithm \cite{lloyd1982least,macqueen1967some}. After the feature selection, we calculated the validation accuracy and plotted it against the number of centers per class in Figure \ref {fig:Isolet_Analysis}. The validation accuracy jumped significantly from one center to two centers per class. The increased accuracy indicates that the speech classes are multi-modal, further validated by the two-dimensional PCA plot of the three classes shown in panel (b)-(d).

\begin{figure}[ht!]
	\centering
	\includegraphics[width=13.5cm,height=9.0cm]{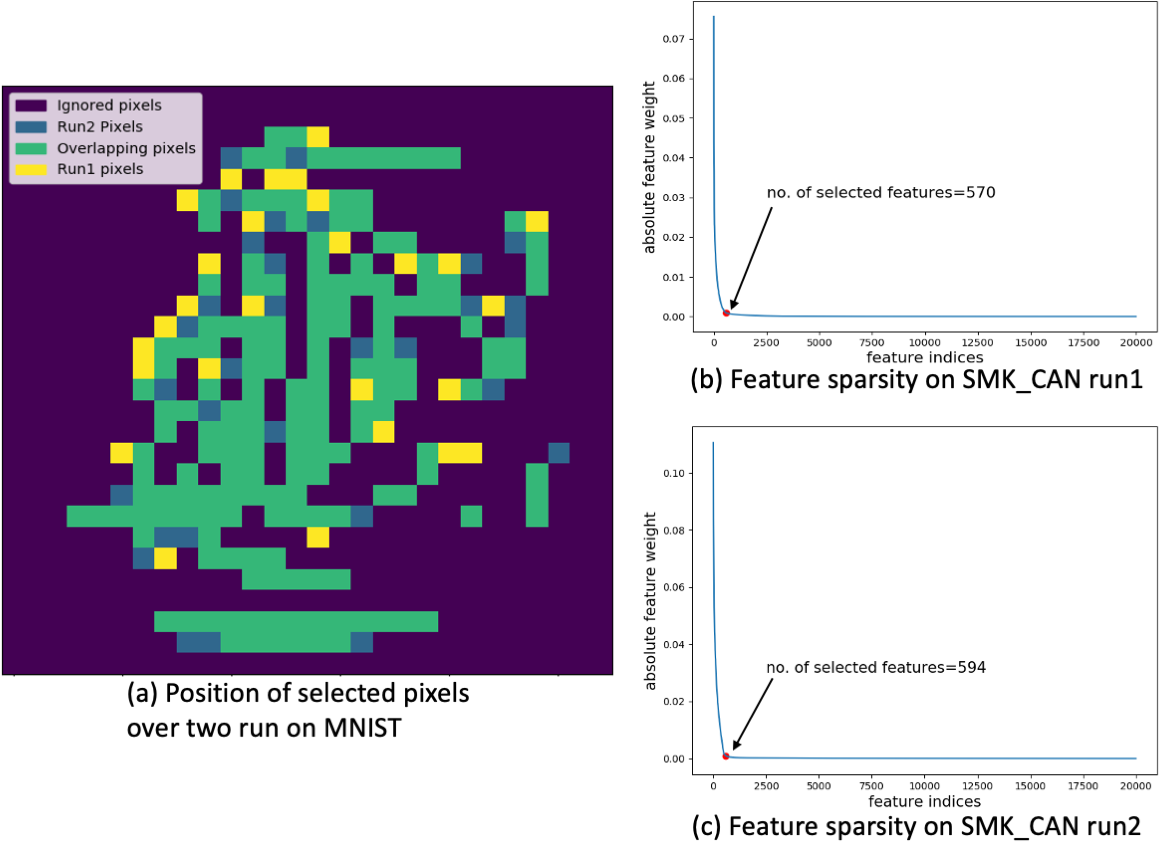}
	\caption{Sparse Centroid-Encoder on MNIST (all ten classes) and high dimensional ($\#$features 19993) SMK\_CAN data. Panel (a) shows position of the selected pixels over two run ($\lambda = 0.0002$). Panel (b)-(c) shows the sparsity plot ($\lambda = 0.0002$) of the SPL layer over two run on SMK\_CAN training data.}
	\label{fig:Feature_Stability}
\end{figure}

Our last analysis sheds light on the feature selection stability of the SCE as shown in Figure \ref{fig:Feature_Stability}. In panel (a), we present the position of the selected pixels over two runs (194 and 198 respectively with 167 overlapping ones). Most of the selected pixels reside in the middle of the image, making sense as the MNIST digits lie in the center of a 28 x 28 grid. Notice that the non-overlapping pixels of the two runs are neighbors, making sense as the neighboring pixels perhaps contain similar information about the digits. On the SMK\_CAN dataset, which has over 19,000 features, the 1-norm of the sparse layer makes most of the variables to zero/near zero, selecting only 570 and 594 biomarkers. We didn't use hard thresholding to induce sparsity. Our feature cut-off technique, mentioned in Section \ref{feature_cut-off}, picks the non-zero biomarkers. Also, notice the absolute weight of the selected biomarkers, which suggests that the $\ell_1$ didn't shrink all parameters of the sparse layers.

\newpage
\section{Experimental Results}
\label{experiments}

We present the comparative evaluation of our model on various data sets using several feature selection techniques. 
%\subsection{Data sets}
%We used five disparate data sets including single-cell data (GM12878), high dimensional infectious disease data set (GSE73072), vision letter data (MNIST), hyperspectral imagery (Indian Pines), and GIS data (Forest Cover).

\subsection{Experimental Details}
We used twelve data sets from a variety of domains (image, biology, speech, and sensor; see Table\ref{table:dataDescription}) and five neural network-based models to run three benchmarking experiments. To this end, we picked the published results from four papers \cite{lemhadri2021lassonet, singh2020fsnet, li2016deep, yamada2020feature} for benchmarking. We followed the same experimental methodology described in those papers for an apples-to-apples comparison. This approach permitted a direct comparison of LassoNet, FsNet, Supervised CAE, DFS, and Stochastic Gates using the authors' best results. All three experiments follow the standard workflow.
\begin{itemize}
    \item  Split each data sets into training and test partition.
    \item Run SCE on the training set to extract top $K\in\{10,16,50\}$ features.
    \item Using the top $K$ features train a one hidden layer ANN classifier with $H$ ReLU units to predict the test samples. The $H$ is picked using a validation set.
    \item Repeat the classification 20 times and report average accuracy.
\end{itemize}
Now we describe the details of the three experiments.

\textbf{Experiment 1:} The first bench-marking experiment is conducted on five real-world high dimensional biological data sets: ALLAML, GLIOMA, SMK\_CAN, Prostate\_GE, and GLI\_85\footnote{Available at http://featureselection.asu.edu/datasets.php} to compare SCE with FsNet and Supervised CAE (SCAE). Following the experimental protocol of Singh et al. \cite{singh2020fsnet}, we randomly partitioned each data into a 50:50 ratio of train and test and ran SCE on the training set. After that, we calculated the test accuracy using the top $K=\{10,50\}$ SCE features. We repeated the experiment 20 times and reported the mean accuracy. We ran a 5-fold cross-validation on the training set to tune the hyperparameters.

\textbf{Experiment 2:} In the second bench-marking experiment, we compared SCE with LassoNet\cite{lemhadri2021lassonet} and Stochastic Gate\cite{yamada2020feature} on six data sets: Mice Protein\footnote{There are some missing entries that are imputed by mean feature values.}, COIL20, Isolet, Human Activity, MNIST, and FMNIST\footnote{Available at UCI Machine Learning repository}. Following the experimental set of Lemhadri et al., we split each data set into 70:10:20 ratio of training, validation, and test sets. We ran SCE on the training set to pick the top $K=50$ features to predict the class labels of the sequester test set. We extensively used the validation set to tune the hyperparameters.

\textbf{Experiment 3:} In the last benchmark, we used the single cell GM12878 data \footnote{The authors of \cite{li2016deep} shared the data with us} which has separate training, validation, and test sets. The SCE is run to select the top $K=16$ features to compare the prediction performance with Deep/Shallow DFS \cite{li2016deep}, and Lasso. Again we used the validation set for hyperparameters tuning.

\begin{table}[!ht]	
	\vspace{1mm}
	\centering
	\begin{tabular} {|c|c|c|c|c|c|}	% (3 columns)
		% start header
		\hline	% Makes 2 fancy lines
		Dataset & No. Features & No. of Classes & No. of Samples & Domain \\
		\hline
		ALLAML & 7129 & 2 & 72 & Biology \\
		GLIOMA & 4434 & 4 & 50 & Biology \\
		SMK\_CAN & 19993 & 2 & 187 & Biology \\
		Prostate\_GE & 5966 & 2 & 102 & Biology \\
		GLI\_85 & 22283 & 2 & 85 & Biology \\
		GM12878 & 93 & 3 & 6468 & Biology \\
		Mice Protein & 77 & 8 & 975 & Biology \\
		COIL20 & 1024 & 20 & 1440 & Image \\
		Isolet & 617 & 26 & 7797 & Speech \\
		Human Activity & 561 & 6 & 5744 & Accelerometer Sensor \\
		MNIST & 784 & 10 & 70000 & Image \\
		FMNIST & 784 & 10 & 70000 & Image \\
		%\hline

		\hline
	\end{tabular}	
	\caption{Descriptions of the data sets used for benchmarking experiments.}
	\label{table:dataDescription}
\end{table}

\subsection{Results}
Now we discuss the results of the three benchmarking experiments. In Table \ref{table:exp1_results} we present the results of the first experiment where we compare SCE, SCAE, and FsNet on five high-dimensional biological data sets. Apart from the results using a subset (10 and 50) of features, we also provide the prediction using all the features. In most cases, feature selection helps improve classification performance. Generally, SCE features perform better than SCAE and FsNet; out of the ten classification tasks, SCE produces the best result on eight. Notice that the top fifty SCE features give a better prediction rate than the top ten in all the cases. Interestingly, the accuracy of SCAE and FsNet drop significantly on SMK\_CAN and GLI\_85 using the top fifty features. 

\begin{table}[h!]
	\vspace{-0.25cm}
	\centering
	\begin{tabular} {|c|c|c|c|c|c|c|c|}	% (3 columns)
		% start header
		\hline 	% Makes 2 fancy lines
		%\multicolumn{1}{|c|}{Method} & \multicolumn{3}{c|} {Dataset}\\
		\multirow{2}{*}{Data set} & \multicolumn{3}{c|} {Top 10 features} & \multicolumn{3}{c|} {Top 50 features}  & \multirow{1}{*}{All features} \\
		\cline{2-7}		
		& \multicolumn{1}{c|} {FsNet} &  \multicolumn{1}{c|} {SCAE} &  \multicolumn{1}{c|} {SCE} & \multicolumn{1}{c|} {FsNet} & \multicolumn{1}{c|} {SCAE} & \multicolumn{1}{c|} {SCE} & \multirow{1}{*}{ANN} \\
		
		\hline
		ALLAML & $91.1$ & $83.3$ & $\textbf{92.5}$ & $92.2$ & $93.6$ & $\textbf{95.9}$ & $89.9$\\
        \hline
        Prostate\_GE & $87.1$ & $83.5$ & $\textbf{89.5}$ & $87.8$ & $88.4$ & $\textbf{89.9}$ & $75.9$\\
		\hline
        GLIOMA & $62.4$ & $58.4$ & $\textbf{63.2}$ & $62.4$ & $60.4$ & $\textbf{69.0}$ & $70.3$\\
		\hline
		SMK\_CAN & $\textbf{69.5}$ & $68.0$ & $68.6$ & $64.1$ & $66.7$ & $\textbf{69.4}$ & $65.7$\\
		\hline
        GLI\_85 & $87.4$ & $\textbf{88.4}$ & $84.1$ & $79.5$ & $82.2$ & $\textbf{85.5}$ & $79.5$\\
        %\hline
        %CLL\_SUB & $\textbf{64.0}$ & $57.5$ & $53.1$ & $\textbf{58.2}$ & $55.6$ & $55.6$ & $56.9$\\
        \hline
		
	\end{tabular}	
	\caption{Comparison of mean classification accuracy of FsNet, SCAE, and SCE features on five real-world high-dimensional biological data sets. The prediction rates are averaged over twenty runs on the test set. Numbers for FsNet and SCAE are being reported from \cite{singh2020fsnet}.}
	\label{table:exp1_results}
\end{table}
Now we turn our attention to the results of the second experiment, as shown in Table\ref{table:exp2_results}. The features of the Sparse Centroid-Encoder produce better classification accuracy than LassoNet and STG in all the cases. Especially for Mice Protein, Activity, Isolet, FMNIST, and MNIST, our model has better accuracy by $2.5\% - 4.5\%$. The results for Stochastic Gates (STG) in \cite{yamada2020feature} are not in a table form, but our eyeball comparison of classification accuracy with the top 50 features on ISOLET, COIL20, and MNIST suggests that stochastic gate is not more accurate than SCE. For example, using the top 50 features, STG obtains approximately 85\% accuracy on ISOLET while SCE obtains 91.1\%; STG obtains about 97\% on COIL20 while SCE obtains 99.3\%; on the dataset, MNIST STG achieves approximately 91\% while SCE 93.8\%. In this experiment, we ran SCE with multiple centroids per class and observed an improved prediction rate than one center per class on Isolet, Activity, MNIST, and FMNIST. The observation suggests that the classes are multi-modal, providing a piece of valuable information. The optimum number of centers was picked using the validation set, see Figure \ref{fig:Isolet_Analysis}.

\begin{table}[ht!]
	\vspace{1mm}
	\centering
	\begin{tabular} {|c|c|c|c|c|c|}	
		% start header
		\hline
		\multirow{2}{*}{Data set} & \multicolumn{3}{c|} {Top 50 features} & \multirow{1}{*}{\#Centers} & \multirow{1}{*}{All features} \\
		\cline{2-4}
		& \multicolumn{1}{c|} {LassoNet} & \multicolumn{1}{c|} {STG} &  \multicolumn{1}{c|} {SCE} & \multicolumn{1}{c|}{for SCE} & {ANN} \\
		
		\hline
		Mice Protein &$95.8$ & NA &$\textbf{98.4}$ & 1 & $100.00$\\
		\hline
		MNIST & $87.3$ & $91.0$ & $\textbf{93.8}$ & 3 & $97.60$\\
		\hline
        FMNIST & $80.0$ & NA &$\textbf{84.7}$ & 3 & $90.16$\\
        \hline
        ISOLET & $88.5$ & $85.0$ &$\textbf{91.1}$ & 5 & $96.96$\\
        \hline
        COIL-20 & $99.1$ & $97.0$ &$\textbf{99.3}$ &  1 & $98.87$\\
        \hline
        Activity & $84.9$ & NA &$\textbf{89.4}$ & 4 & $92.81$\\
		\hline
	\end{tabular}	
	\caption{Classification results using LassoNet, STG, and SCE features on six publicly available data sets. The column '\#Centers for SCE' denotes how many centroids per class are used to train SCE. Numbers for LassoNet and STG are reported from \cite{lemhadri2021lassonet} and \cite{yamada2020feature} respectively. All the reported accuracies are measured on the test set. NA means the result has not been reported.}
	\label{table:exp2_results}
\end{table}

In Table\ref{table:exp3_results}, we present the results of our last experiment on the single cell data GM12878. We use the published results for deep feature selection (DFS), shallow feature selection, and Lasso from the work of Li et al. to evaluate SCE. To compare with Li et al., we used the top 16 features to report the mean accuracy of the test samples. We see that the SCE features outperform all the other models. Among all the models, Lasso exhibits the worst performance with an accuracy of $81.86\%$. This relatively low accuracy is not surprising, given Lasso is a linear model.

\begin{table}[ht!]
	%\vspace{1mm}
	\centering
	\begin{tabular} {|c|c|c|c|c|}	
		% start header
		\hline
		\multirow{2}{*}{Data set} & \multicolumn{4}{c|} {Top 16 features} \\
		\cline{2-5}
		& \multicolumn{1}{c|} {SCE} & \multicolumn{1}{c|} {Deep DFS} &  \multicolumn{1}{c|} {Shallow DFS} & \multicolumn{1}{c|}{Lasso} \\
		
		\hline
		GM12878 & \textbf{87.51} & 85.67 & 85.34 & 81.86 \\

		\hline
	\end{tabular}	
	\caption{Classification accuracies using the top 16 features by various techniques. Results of Deep DFS, Shallow DFS, Lasso, and Random Forest are reported from \cite{li2016deep}.}
	\label{table:exp3_results}
\end{table}

The classification performance gives a quantitative measure that doesn't reveal the biological significance of the selected genes. We did a literature survey of the top genes selected by sparse centroid-encoder and provided a detailed description in the appendix. Some of these genes play an essential role in transcriptional activation, e.g., H4K20ME1 \cite{barski2007high}, TAF1 \cite{wang2014crystal}, H3K27ME3 \cite{cai2021h3k27me3}, etc. Gene H3K27AC \cite{creyghton2010histone} plays a vital role in separating active enhances from inactive ones. Besides that,  many of these genes are related to the proliferation of the lymphoblastoid cancer cells, e.g., POL2 \cite{yamada2013nuclear}, NRSF/REST \cite{kreisler2010regulation}, GCN5 \cite{yin2015histone}, PML \cite{salomoni2002role}, etc. This survey confirms the biological significance of the selected genes.

\section{Related Work}
\label{lit}

Feature selection has a long history spread across many fields, including bioinformatics, document classification, data mining, hyperspectral band selection, computer vision, etc. It's an active research area, and numerous techniques exist to accomplish the task. We describe the literature related to the embedded methods where the selection criteria are part of a model. The model can be either linear or non-linear.

\subsection{Feature Selection using Linear Models}
%Linear models are widely used in Machine Learning for classification and regression. These models approximate the output as a linear combination of input variables (features), i.e. $y \approx f(x) = w^T x + b$ where $w$ and $b$ are the model parameters. From the optimization perspective, a linear models takes the following form:
%$\underset{\theta}{minimize}\;\; l(y,f(x,\theta))$
%where $l$ is a loss function and $\theta$ is the parameter set. 

Adding an $l_1$ penalty to classification and regression methods naturally produce feature selectors.
%on the parameter set $\theta$ gives a feature selector, 
For example, least absolute shrinkage and selection operator or Lasso \cite{tibshirani1996regression}
%, which takes the form: 
%\begin{equation}
%    \begin{aligned}
%\underset{\theta}{minimize}\;\; l(y,f(x,\theta)) \; + \; %\lambda\| \theta \|_1 
%    \end{aligned}
%\end{equation}
%where $\lambda$ is a hyper parameter which controls the sparsity of the model. Since its inception, the model 
has been used extensively for feature selection on various data sets \cite{fonti2017feature,muthukrishnan2016lasso,kim2004gradient}.
Elastic net, proposed by Zou et al. \cite{zou2005regularization}, combined the Lasso penalty with the Ridge Regression penalty \cite{hoerl1970ridge} to overcome some limitations of Lasso. 
%The Elastic net is defined as following: 
%\begin{equation}
%    \begin{aligned}
%\underset{\theta}{minimize}\;\; l(y,f(x,\theta)) \; + \; (1-\alpha)\| \theta \|_1 \;+ \; \alpha \| \theta \|^2_2 
%    \end{aligned}
%\end{equation}
%where $\alpha \in [0,1)$ and the term $(1-\alpha)\| \theta \|_1 +  \alpha \| \theta \|^2_2$ known as elastic net penalty. 
Elastic net has been widely applied, e.g.,  \cite{marafino2015efficient,shen2011identifying,sokolov2016pathway}. Note both Lasso and Elastic net are convex in the parameter space. Also see the following works using $\ell_1$ based features selection on linear models see \cite{lindenbaum2021randomly,candes2008enhancing,daubechies2010iteratively,bertsimas2017trimmed,xie2009scad}.
Support Vector Machines (SVM) \cite{cortes1995support} is a state-of-the-art model for classification, regression and feature selection. SVM-RFE is a linear feature selection model which iteratively removes the least discriminative features until a parsimonious set of predictive features are selected \cite{guyon2002gene}. IFR \cite{o2013iterative}, on the other hand, selects a group of discriminatory features at each iteration and eliminates them from the data set. The process repeats until the accuracy of the model starts to drop significantly. Note IFR uses Sparse SVM (SSVM), which minimizes the $l_1$ norm of the model parameters. Lasso, Elastic Net, and SVM-based techniques are mainly applied to binary problems. These models are extended to the multi-class problem by combining multiple binary one-against-one (OAO) or one-against-all (OAA) models.  \cite{chepushtanova2014band} used 120 Sparse SVM models to select discriminative bands from the Indian Pine data set, which has 16 classes.
On the other hand, Random forest \cite{breiman2001random}, a decision tree-based technique, finds features from multi-class data using a single model. The model doesn't use Lasso or Elastic net penalty for feature selection. Instead, the model weighs the importance of each feature by measuring the out-of-bag error.

\subsection{Feature Selection using Deep Neural Networks}
While the linear models are fast and convex, they don't capture the non-linear relationship among the input features (unless a kernel trick is applied). Because of the shallow architecture, these models don't learn a high-level representation of input features. Moreover, there is no natural way to incorporate multi-class data in a single model. Non-linear models based on deep neural networks overcome these limitations. In this section, we will briefly discuss a handful of such models.

\cite{scardapane2017group} used group Lasso \cite{tibshirani1996regression} to impose the sparsity on a group of variables instead of a single variable. They applied the group sparsity simultaneously on the input and the hidden layers to remove features from the input data and the hidden activation. 
%On MNIST, their algorithm discarded more than 200 features from the input vector with an accuracy of $97\%$ on the test data. Although on the Forest Cover data set, the algorithm used most of the input variables $52.7$ out of $54$.
Li et al. proposed deep feature selection (DFS), which is a multilayer neural network-based feature selection technique \cite{li2016deep}. DFS uses a one-to-one linear layer between the input and the first hidden layer. As a sparse regularization, the authors used elastic-net \cite{zou2005regularization} on the variables of the one-to-one layer to induce sparsity. The standard soft-max function is used in the output layer for classification. With this setup, the network is trained in an end-to-end fashion by error backpropagation. Despite the deep architecture, its accuracy is not competitive, and experimental results have shown that the method did not outperform the random forest (RF) method. \cite{kim2016opening} proposed a heuristics based technique to assign importance to each feature. Using the ReLU activation, \cite{roy2015feature} provided a way to measure the contribution of an input feature towards hidden activation of next layer.
%{\color{red} \cite{kim2016opening} proposed a heuristics based technique to assign importance to each feature. Using the ReLU activation, \cite{roy2015feature} provided a way to measure the contribution of an input feature towards hidden activation of next layer.}
%{\color{blue} Looks good, not suggested changes.}
\cite{han2018autoencoder} developed an unsupervised feature selection technique based on the autoencoder architecture. Using a $l_{2,1}$-norm to the weights emanating from each input node, they measure the contribution of each feature while reconstructing the input. The model excavates the input features, which have a minimum contribution.
\cite{taherkhani2018deep} proposed a RBM \cite{Hinton:2006:FLA:1161603.1161605,hinton2006reducing} based feature selection model. 
%In this approach, an RBM is trained using the input data. Then, the reconstruction error ($e_i$) of each $i^{th}$ feature is calculated. After that, the value of the $i^{th}$ input node is set to zero, and the reconstruction error ($\tilde e_{i}$) of the same input node is measured again. If  $\tilde e_{i} <= e_i$, i.e. the $i^{th}$ feature can be reconstructed from the other features and hence the feature is discarded. In the implementation, the authors randomly select the $N_{c}$ number of features to run the algorithm. 
This algorithm runs the risk of combinatorial explosion for data set with $50K-60K$ features (e.g., microarray gene expression data set).

\section{Discussion and Conclusion}
\label{dis_cons}

In this paper, we presented Sparse Centroid-Encoder as an efficient feature selection tool for binary and multiclass problems. The benchmarking results span twelve diverse data sets and six methods providing evidence that the features of SCE produce better generalization performance than other state-of-the-art models. We compared SCE with FsNet, mainly designed for high-dimensional biological data, and found that our proposed method outperformed it in most cases. The comparison also includes Supervised CAE, which is not more accurate than SCE. On the data sets, where the no. of observations are more than the no. of variables, SCE produces the best classification results than LassoNet, Stochastic Gate, and DFS in each case. These experiments involve image, speech, and accelerometer sensor data. Moreover, the survey of the sixteen SCE genes of GM12878 indicates plausible biological significance. The empirical evaluation using an array of diverse data sets establishes the value of the Sparse Centroid-Encoder as a nonlinear feature detector.

Our analysis of the Sparse Centroid-Encoder in Section \ref{sparsity_analysis} demonstrates that the 1-norm induces good feature sparsity. We chose the $\lambda$ from the validation set from a wide range of values and saw that smaller values work better for classification. The visualization of the MNIST pixels (panel (a) of Figure \ref{fig:Feature_Stability}) provides a qualitative justification for a high prediction rate. SCE selected most of the pixels from the central part of the image, ignoring the border, making sense as the digits lie in the center of a 28 x 28 grid. The $\ell_1$ penalty on the SPL layer induces sharp sparsity on the SMK\_CAN data (panel (b) and (c) of Figure \ref{fig:Feature_Stability}) without shrinking all the variables. Our feature cut-off technique correctly demarcates the important features from the rest.

SCE compares favorably to the neural network-based model FsNet, SCAE DFS, LassoNet, and STG, where samples are mapped to class labels. SCE employs multiple centroids to capture the variability within a class, improving the prediction rate of unknown test samples. In particular, the prediction rate on the ISOLET improved significantly from one centroid to multiple centroids suggesting the speech classes are multi-modal. The two-dimensional PCA of ISOLET classes further confirms the multi-modality of data. We also observed an enhanced classification rate on MNIST, FMNIST, and Activity data with multiple centroids. 

In contrast, single-center per class performed better for other data sets (e.g., COIL-20, Mice Protein, GM12878, etc.). Hence, apart from producing an improved prediction rate using features that capture intra-class variance, our model can provide extra information about whether the data is unimodal or multi-modal. This aspect of sparse centroid-encoder distinguishes it from the techniques which do not model the multi-modal nature of the data.

\bibliographystyle{unsrt}  
\bibliography{SCE_arXiv}

\end{document}